\documentclass{bmvc2k}

\usepackage{times}
\usepackage{epsfig}
\usepackage{graphicx}
\usepackage{amsmath}
\usepackage{amssymb}
\usepackage{color}
\usepackage{comment}
\usepackage[normalem]{ulem}
\usepackage{caption}
\usepackage{booktabs}
\usepackage{multirow}
\usepackage{subfig}
\usepackage[export]{adjustbox}
\graphicspath{{figures/}}

\definecolor{darkgreen}{RGB}{0,127,0}
\definecolor{darkblue}{RGB}{0,0,127}
\definecolor{darkred}{RGB}{127,0,0}
\definecolor{darkmagenta}{RGB}{127,0,127}
\definecolor{darkcyan}{RGB}{0,127,127}
\definecolor{darkyellow}{cmyk}{0,0.7,0.9,0.3}

\renewcommand{\paragraph}[1]{{\vspace{0mm}\flushleft{\textbf{#1}}}} 
\long\def\ignorethis#1{}

\usepackage[pagebackref=true,urlcolor=blue,citecolor=red,breaklinks=true,letterpaper=true,colorlinks,bookmarks=false]{hyperref}

\title{Video Stitching for Linear Camera Arrays}

\addauthor{Wei-Sheng Lai$^{1,}$}{wlai24@ucmerced.edu}{2}
\addauthor{Orazio Gallo}{ogallo@nvidia.com}{2}
\addauthor{Jinwei Gu}{gujinwei@gmail.com}{2}
\addauthor{Deqing Sun}{deqing.sun@gmail.com}{2}
\addauthor{Ming-Hsuan Yang}{mhyang@ucmerced.edu}{1}
\addauthor{Jan Kautz}{jkautz@nvidia.com}{2}

\addinstitution{University of California, Merced}
\addinstitution{NVIDIA}

\runninghead{Lai et al.}{Video Stitching for Linear Camera Arrays}

\def\eg{\emph{e.g}\bmvaOneDot}
\def\ie{\emph{i.e}\bmvaOneDot}

\def\etal{\emph{et al}\bmvaOneDot}

\begin{document}

\maketitle

\begin{figure}
\vspace{-5mm}
\begin{center}

\newcommand{\teaserHratio}{0.22}

\subfloat[Our stitching result]{\includegraphics[height=\teaserHratio\textwidth]{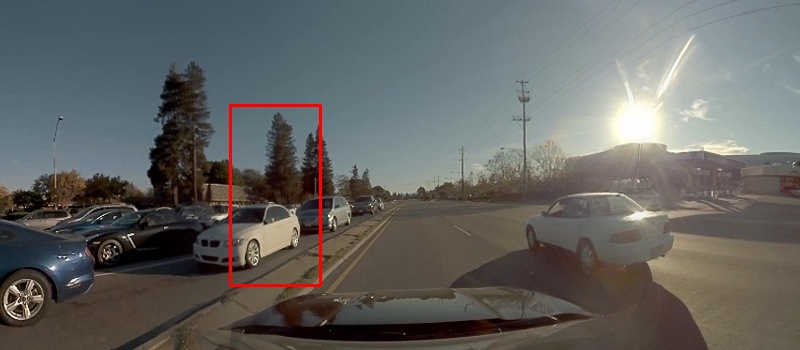}}~
\subfloat[\cite{vrworks}]{\includegraphics[height=\teaserHratio\textwidth]{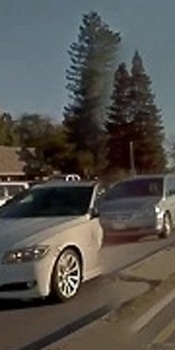}}~
\subfloat[\cite{autopano}]{\includegraphics[height=\teaserHratio\textwidth]{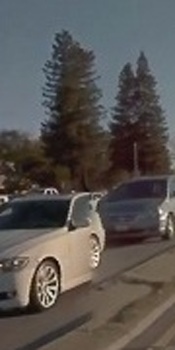}}~
\subfloat[\cite{jiang2015video}]{\includegraphics[height=\teaserHratio\textwidth]{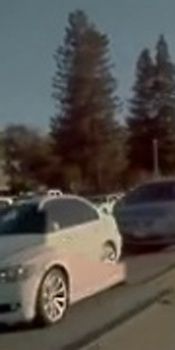}}~
\subfloat[Ours]{\includegraphics[height=\teaserHratio\textwidth]{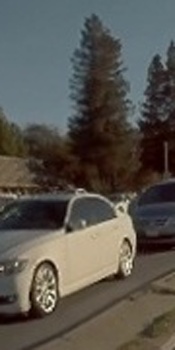}}

\end{center}
\caption{
\textbf{Examples of video stitching.} 
Inspired by pushbroom cameras, we propose a deep pushbroom stitching network to stitch multiple wide-baseline videos of dynamic scenes into a single panoramic video.
The proposed learning-based algorithm outperforms prior work with minimal mis-alignment artifacts (\eg, ghosting and broken objects).
More video results are presented in the supplementary material.
}\label{fig:teaser}
\end{figure}
\vspace{-5mm}

\begin{abstract}
Despite the long history of image and video stitching research, existing academic and commercial solutions still produce strong artifacts.
In this work, we propose a wide-baseline video stitching algorithm for linear camera arrays that is temporally stable and tolerant to strong parallax. 
Our key insight is that stitching can be cast as a problem of learning a smooth spatial interpolation between the input videos.
To solve this problem, inspired by pushbroom cameras, we introduce a fast pushbroom interpolation layer and propose a novel pushbroom stitching network, which learns a dense flow field to smoothly align the multiple input videos for spatial interpolation. 
Our approach outperforms the state-of-the-art by a significant margin, as we show with a user study, and has immediate applications in many areas such as virtual reality, immersive telepresence, autonomous driving, and video surveillance.	
	

\end{abstract}	
	
\section{Introduction}
\label{sec:intro}

Due to sensor resolution and optics limitations, the field of view (FOV) of most cameras is too narrow for applications such as autonomous driving and virtual reality. 
A common solution is to stitch the outputs of multiple cameras into a panoramic video, effectively extending the FOV.
When the optical centers of these cameras are nearly co-located, stitching can be solved with a simple homography transformation.
However, in many applications, such as autonomous driving, remote video conference, and video surveillance, multiple cameras have to be placed with \emph{wide baselines}, either to increase view coverage or due to some physical constraints.  
In these cases, even state-of-the-art methods~\cite{jiang2015video, perazzi2015panoramic} and current commercial solutions (\eg, VideoStitch Studio~\cite{video_stitch_studio}, AutoPano Video~\cite{autopano}, and NVIDIA VRWorks~\cite{vrworks}) struggle to produce artifact-free videos, as shown in Figure~\ref{fig:teaser}. 

One main challenge for video stitching with wide baselines is \emph{parallax}, \ie the apparent displacement of an object in multiple input videos due to camera translation. 
Parallax varies with object depth, which makes it impossible to properly align objects without knowing dense 3D information.
In addition, occlusions, dis-occlusions, and limited overlap between the FOVs also cause a significant amount of stitching artifacts.
To obtain better alignment, existing image stitching algorithms perform content-aware local warping~\cite{chang2014shape,zaragoza2013projective} or find optimal seams around objects to mitigate artifacts at the transition from one view to the other~\cite{eden2006seamless,zhang2014parallax}.
Applying these strategies to process a video frame-by-frame inevitably produces noticeable jittering or wobbling artifacts.
On the other hand, algorithms that explicitly enforce temporal consistency, such as spatio-temporal mesh warping with a large-scale optimization~\cite{jiang2015video}, are computationally expensive.
In fact, commercial video stitching software often adopts simple seam cutting and multi-band blending~\cite{burt1983multiresolution}.
These methods, however, often cause severe artifacts, such as ghosting or misalignment, as shown in Figure~\ref{fig:teaser}.
Moreover, seams can cause objects to be cut off or completely disappear from stitched images---a particularly dangerous outcome for use cases such as autonomous driving.

\begin{figure*}
	\centering
	\footnotesize
	\includegraphics[width=1.0\textwidth]{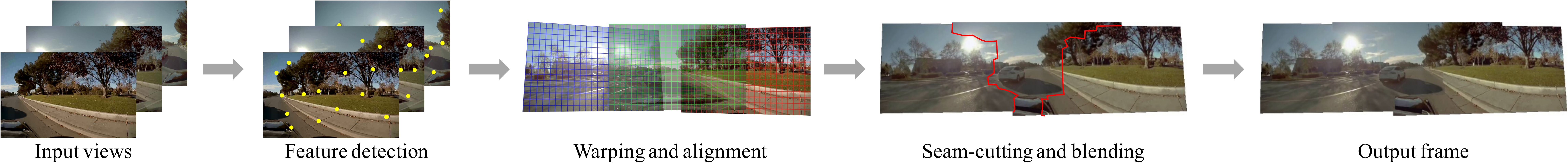}
	\\
	(a) Conventional video stitching pipeline~\cite{jiang2015video}
	\\
	\includegraphics[width=0.65\textwidth]{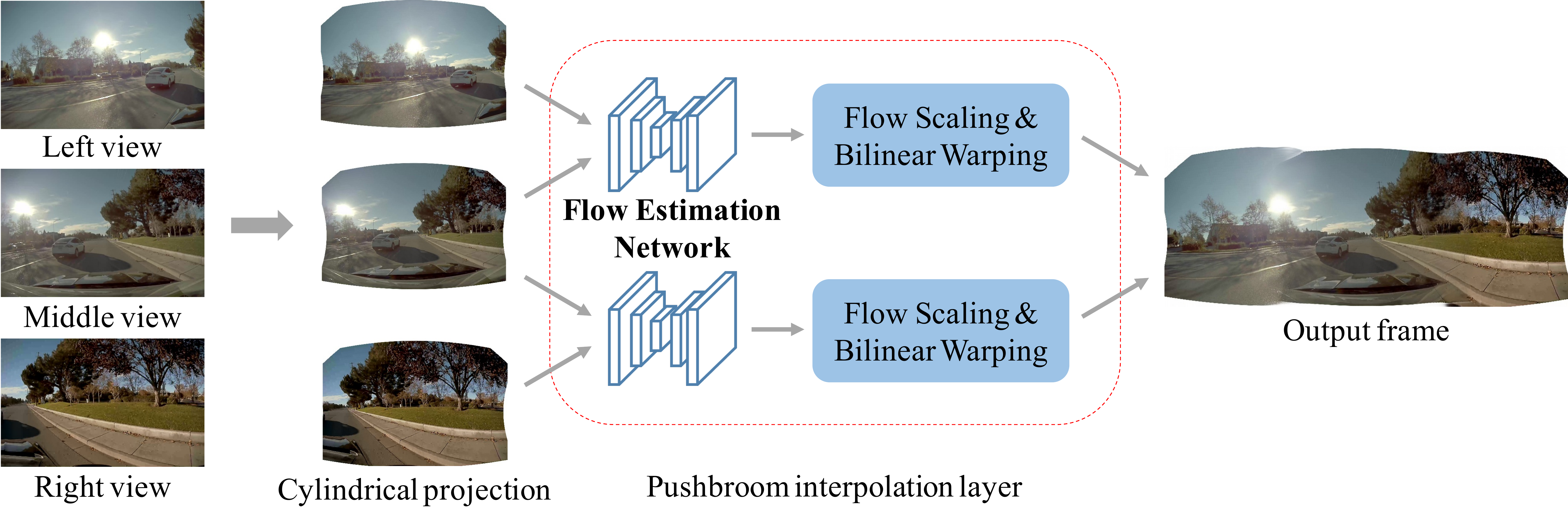}
	\\
	(b) Proposed pushbroom stitching network
	\vspace{-1mm}
	\caption{
		\textbf{Algorithm overview.}
		(a) Conventional video stitching algorithms~\cite{jiang2015video} use spatio-temporal local mesh warping and 3D graph cut to align the entire video, which are often sensitive to scene content and computationally expensive.
		(b) The proposed pushbroom stitching network adopts a pushbroom interpolation layer to gradually align the input views, and outperforms prior work and commercial software.
	}
	\label{fig:overview}
	\vspace{-2mm}
\end{figure*}

We propose a video stitching solution for linear cameras arrays that produces a panoramic video.
We identify three desirable properties in the output video: 
(1) Artifacts, such as ghosting, should not appear. 
(2) Objects may be distorted, but should not be cut off or disappear in any frame. 
(3) The stitched video needs to be temporally stable.
With these three desiderata in mind, we formulate video stitching as a spatial view interpolation problem.
Specifically, we take inspiration from the pushbroom camera, which concatenates vertical image slices that are captured while the camera translates~\cite{gupta1997linear}.
%
We propose a pushbroom stitching network based on deep convolutional neural networks (CNNs).
Specifically, we first project the input views onto a common cylindrical surface.
We then estimate bi-directional optical flow, with which we \emph{simulate} a pushbroom camera by interpolating all the intermediate views between the input views.
Instead of generating all the intermediate views (which requires multiple bilinear warping steps on the entire image), we develop a \emph{pushbroom interpolation layer} to generate the interpolated view in a single feed-forward pass.
Figure~\ref{fig:overview} shows an overview of the conventional video stitching pipeline and our proposed method.
Our method yields results that are visually superior to existing solutions, both academic and commercial, as we show with an extensive user study.

\section{Related Work}
\label{sec:related}

\paragraph{Image stitching.}
Existing image stitching methods often build on the conventional pipeline of Brown and Lowe~\cite{brown2007automatic}, which first estimates a 2D transformation (\eg, homography) for alignment and then stitches the images by defining seams~\cite{eden2006seamless} and using multi-band blending~\cite{burt1983multiresolution}.
However, ghosting artifacts and mis-alignment still exist, especially when input images have large parallax.
To account for parallax, several methods adopt spatially varying local warping based on the affine~\cite{lin2011smoothly} or projective~\cite{zaragoza2013projective} transformations.
Zhang~\etal~\cite{zhang2014parallax} integrate the content-preserving warping and seam-cutting algorithms to handle parallax while avoiding local distortions.
More recent methods combine the homography and similarity transforms~\cite{chang2014shape,lin2015adaptive} to reduce the projective distortion (\ie, stretched shapes) or adopt a global similarity prior~\cite{chen2016natural} to preserve the global shape of the resulting stitched images.

While these methods perform well on still images, applying them to videos frame-by-frame results in strong temporal instability.
In contrast, our algorithm, also single-frame, generates videos that are spatio-temporally coherent, because our pushbroom layer only operates on the overlapping regions, while the rest is directly taken from the inputs.

\paragraph{Video stitching.}
Due to computational efficiency, it is not straightforward to enforce spatio-temporal consistency in existing image stitching algorithms.
Commercial software, \eg, VideoStitch Studio~\cite{video_stitch_studio} or AutoPano Video~\cite{autopano}, often finds a fixed transformation (with camera calibration) to align all the frames, but cannot align local content well.
Recent methods integrate local warping and optical flow~\cite{perazzi2015panoramic} or find a spatio-temporal content-preserving warping~\cite{jiang2015video} to stitch videos, which are computationally expensive.
Lin~\etal~\cite{lin2016seamless} stitch videos captured from hand-held cameras based on 3D scene reconstruction, which is also time-consuming. 
On the other hand, several approaches, \eg, Rich360~\cite{lee2016rich360} and Google Jump~\cite{anderson2016jump}, create $360^\circ$ videos from multiple videos captured on a structured rig.
Recently, NVIDIA released VRWorks~\cite{vrworks}, a toolkit to efficiently stitch videos based on depth and motion estimation.
Still, as shown in Figure~\ref{fig:teaser}(b), several artifacts, \eg, broken objects and ghosting, are visible in the stitched video.

In contrast to existing video stitching methods, our algorithm learns local warping flow fields based on a deep CNN to effectively and efficiently align the input views. The flow is learned to optimize the quality of the stitched video in an end-to-end fashion.
%

\paragraph{Pushbroom panorama.}
Linear pushbroom cameras~\cite{gupta1997linear} are common for satellite imagery: while a satellite moves along its orbit, they capture multiple 1D images, which can be concatenated into a full image.
A similar approach has also been used to capture street view images~\cite{seitz2003multiperspective}.
However, when the scene is not planar, or cannot be approximated as such, they introduce artifacts, such as stretched or compressed objects.
Several methods handle this issue by estimating scene depth~\cite{rav2004mosaicing}, finding a cutting-seam on the space-time volume~\cite{wexler2005space}, or optimizing the viewpoint for each pixel~\cite{agarwala2006photographing}.
The proposed method is a software simulation of a pushbroom camera which creates panoramas by concatenating vertical slices that are spatially interpolated between the input views.
We note that the method of Jin~\etal~\cite{jin2018learning} addresses a similar problem of view morphing, which aims at synthesizing intermediate views along a circular path.
However, they focus on synthesizing a single object, \eg a person or a car, and do not consider the background.
Instead, our method synthesizes intermediate views for the entire scene.

\newcommand{\ci}{C^i}
\newcommand{\co}{C^o}

\section{Stitching as Spatial Interpolation}
\label{sec:algorithm}

Our method produces a temporally stable stitched video from wide-baseline inputs of dynamic scenes.
While the proposed approach is suitable for a generic linear camera array configuration, here we describe it with reference to the automotive use case.
Unlike other applications of structured camera arrays, in the automotive case, objects can come arbitrarily close to the cameras, thus requiring the stitching algorithm to tolerate large parallax.

\begin{figure}
	\centering
	\footnotesize
	\subfloat[Camera setup]{\includegraphics[width=0.5\columnwidth, trim={85 175 85 180}, clip]{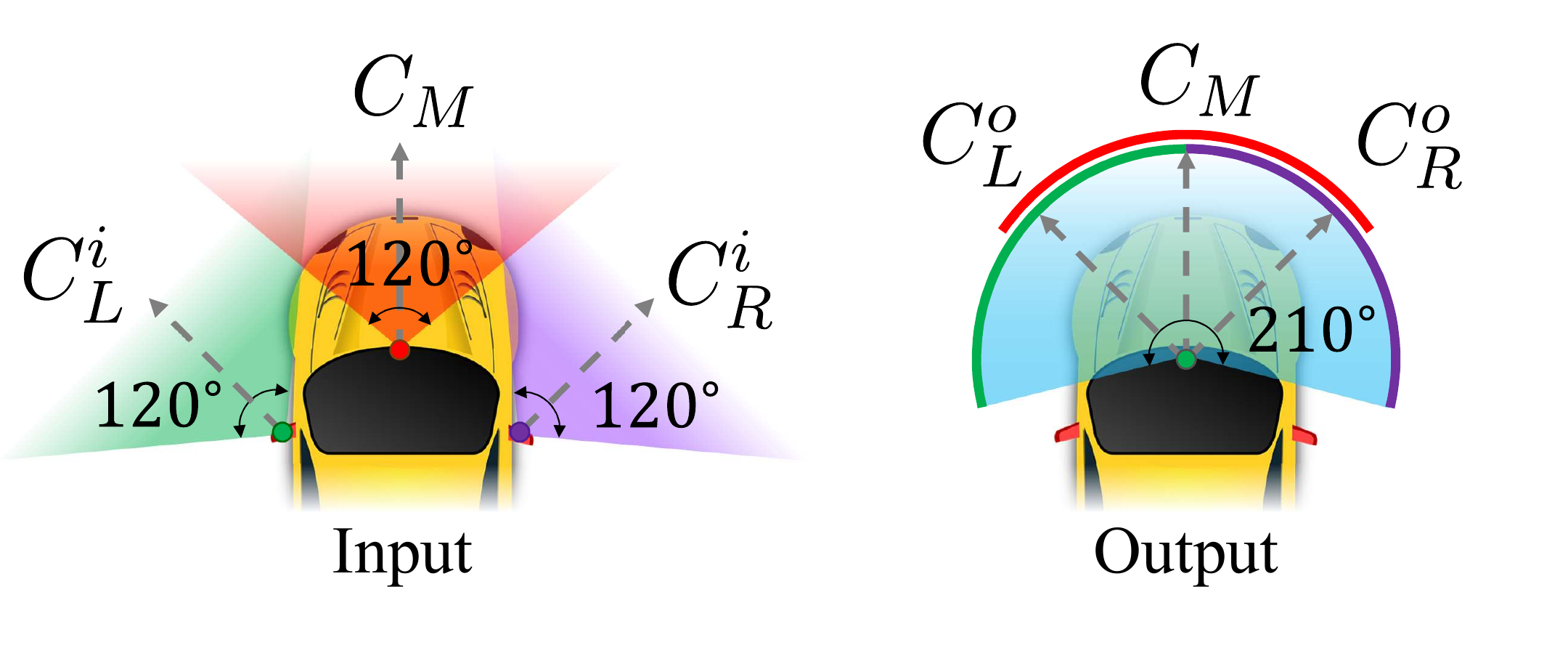}}
	\subfloat[Images on the viewing cylinder]{\includegraphics[width=0.5\columnwidth]{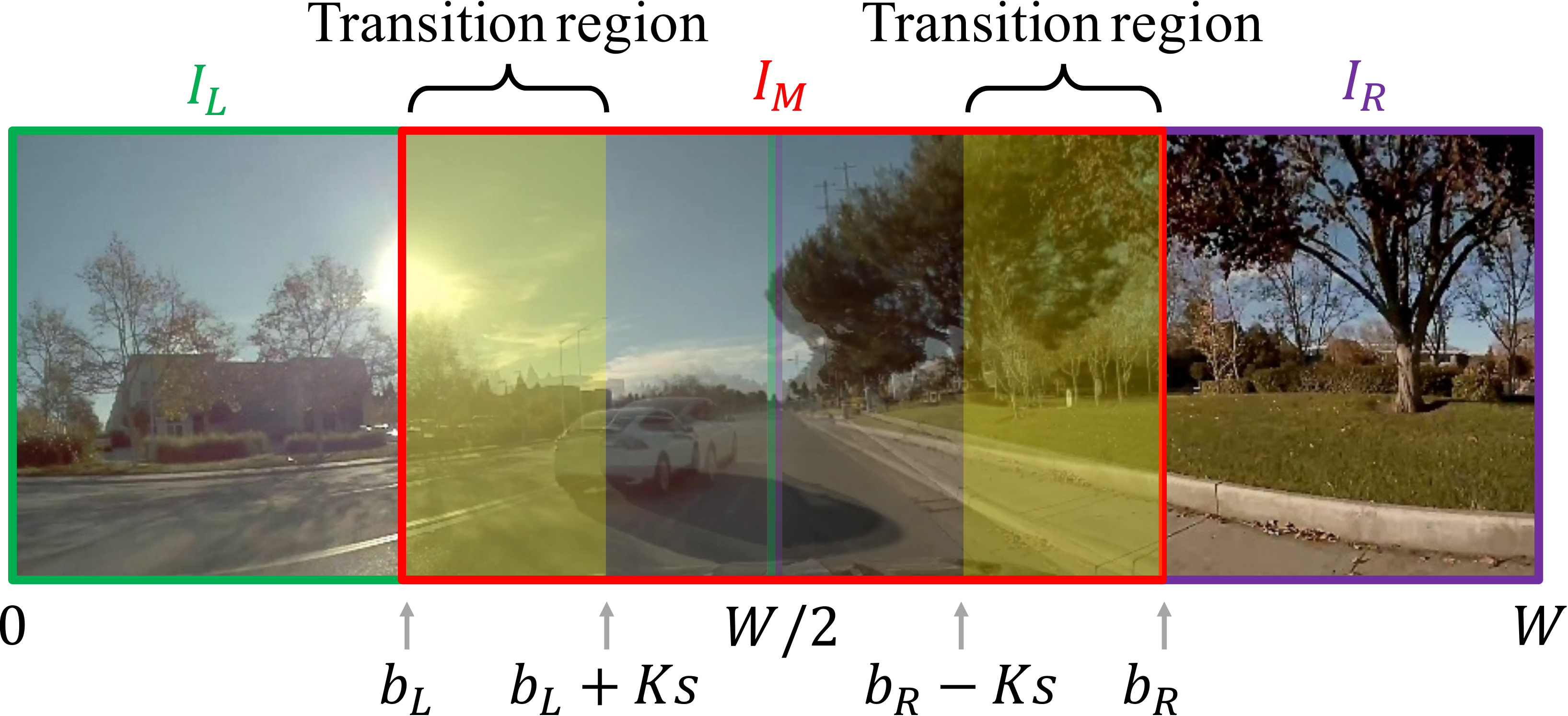}}
	\vspace{2mm}
	\caption{
		\textbf{Camera setup and input images.}
		(a) The input videos are captured from three fisheye cameras, $\ci_L$, $C_{M}$, and $\ci_R$, separated by a large baseline. The output is a viewing cylinder centered at $C_{M}$.
		(b) The input images are first projected on the output cylinder assuming a constant depth. Within the transition regions, our pushbroom interpolation method progressively warps and blends $K$ vertical slices from the input views to create a smooth transition.
		Outside the transition regions, we do not modify the content from the inputs.
	}
	\label{fig:camera}
\end{figure}

\begin{figure}
	\centering
	\subfloat[$I_{L}$]{\includegraphics[width=0.295\columnwidth,valign=m]{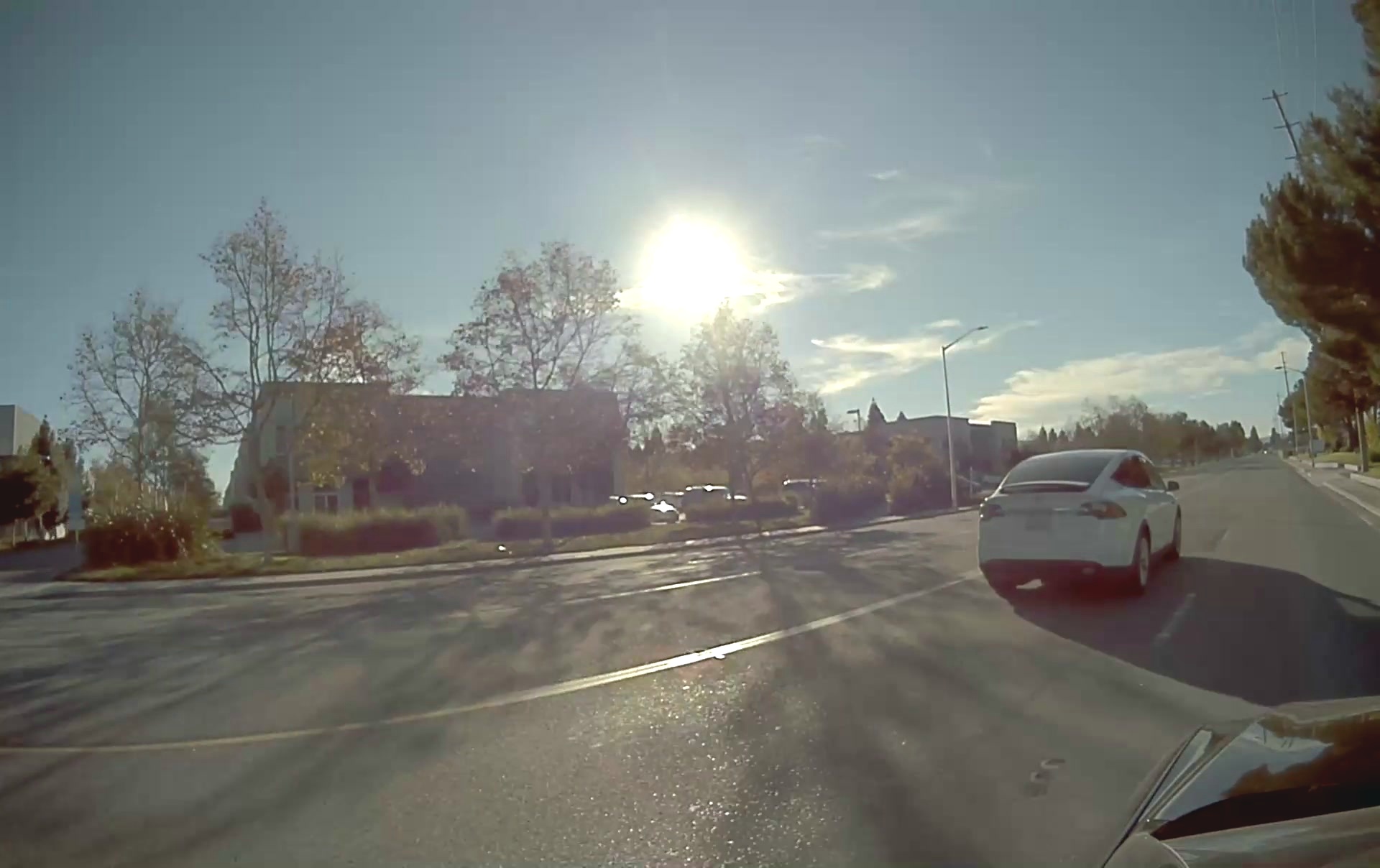} }
	\subfloat[$I_{M}$]{\includegraphics[width=0.295\columnwidth,valign=m]{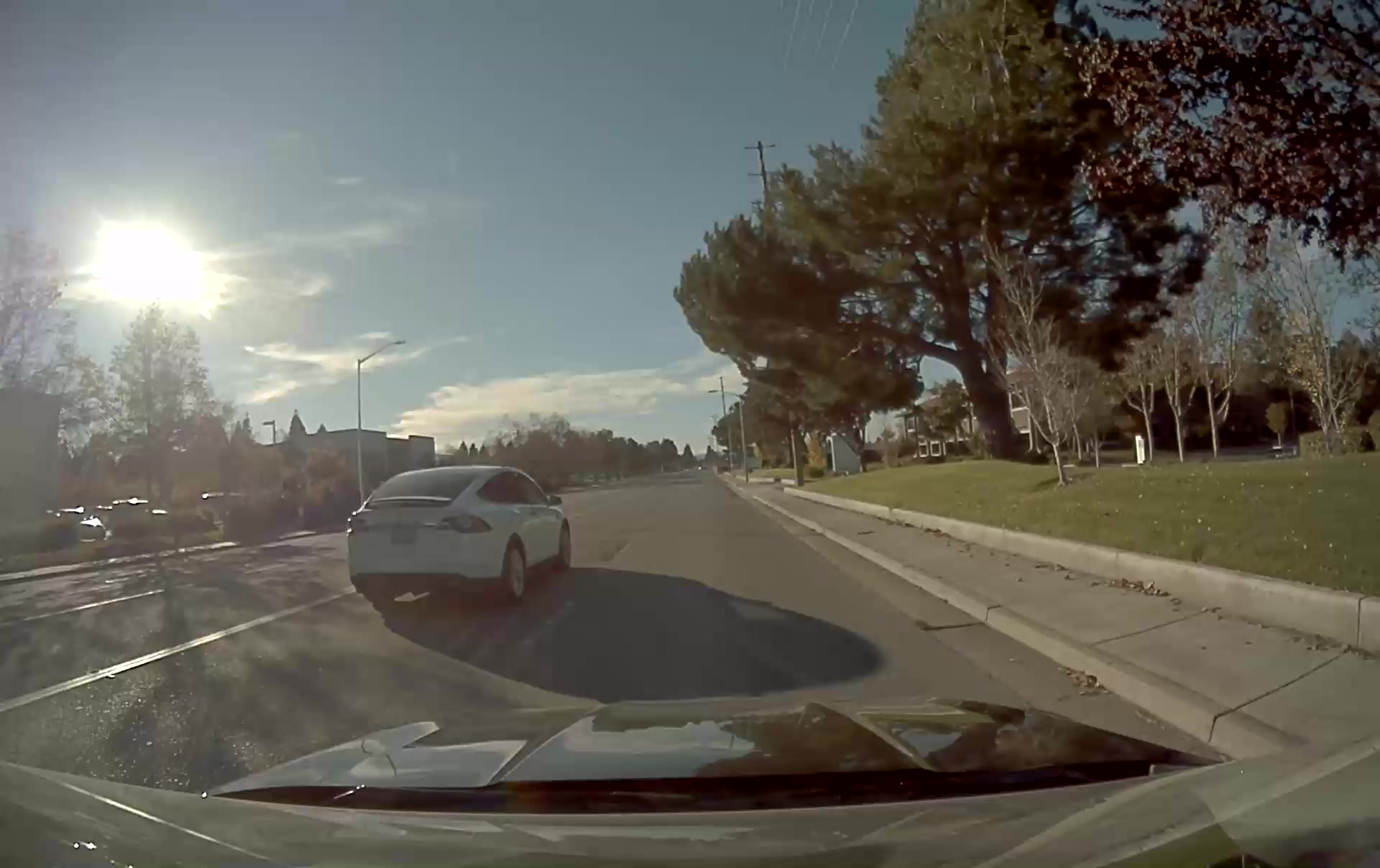} }
	\subfloat[$I_{R}$]{\includegraphics[width=0.295\columnwidth,valign=m]{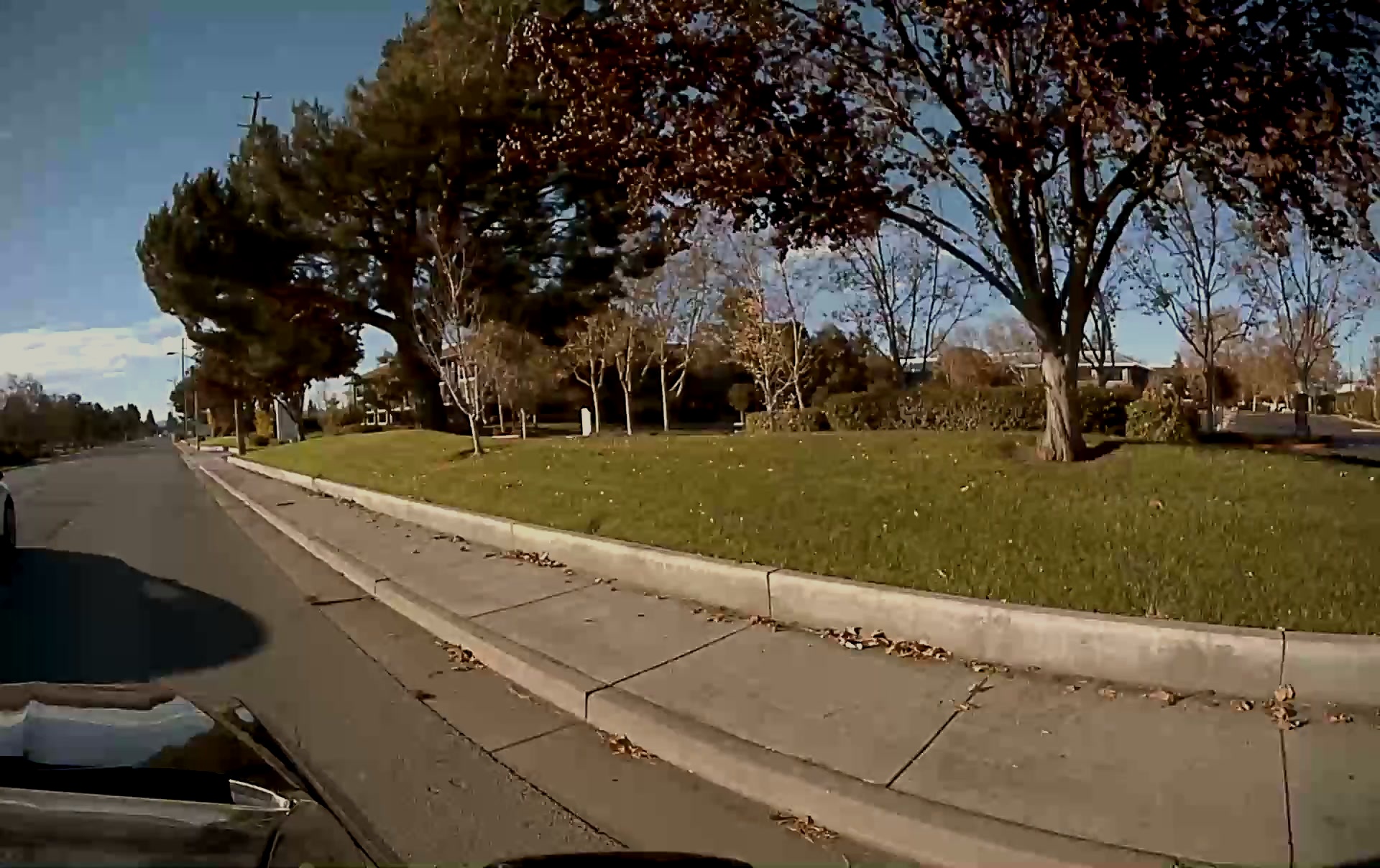} }
	\\
	\subfloat[Images blended on the cylinder]{\includegraphics[width=0.45\columnwidth,valign=m]{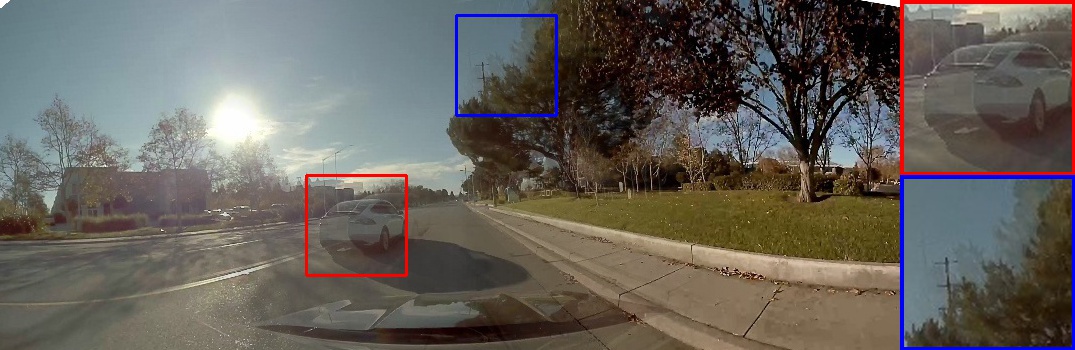} }
	\subfloat[Pushbroom interpolation]{\includegraphics[width=0.45\columnwidth,valign=m]{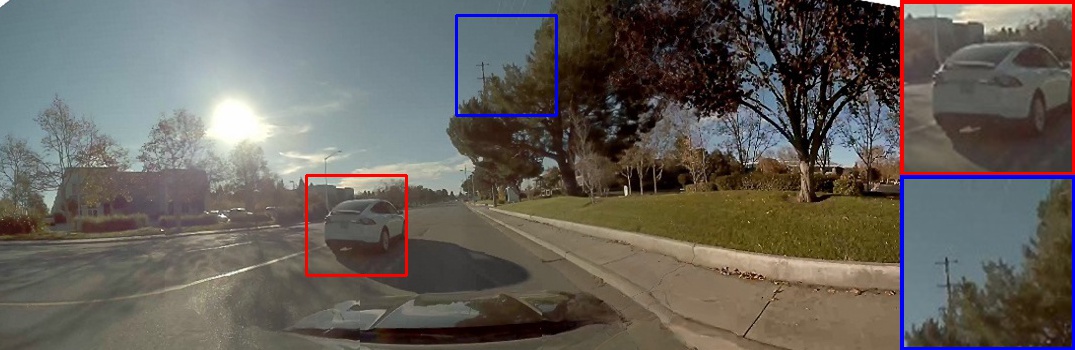} }
	\vspace{2mm}
	\caption{
		\textbf{Example of input and stitched views.}
		Simply projecting the input views $I_L$, $I_M$, and $I_R$ onto the output cylinder causes artifacts due to the parallax and scene depth variation (d).
		Our pushbroom interpolation method effectively stitches the views and does not produce ghosting artifacts (e).
	}
	\label{fig:cylinder}
\end{figure}

For the purpose of describing the method, we define the camera setup as shown in Figure~\ref{fig:camera}(a), which consists of three fisheye cameras whose baseline spans the entire car's width.
Figures~\ref{fig:cylinder}(a)-(c) show typical images captured under this configuration, and underscore some of the challenges we face: strong parallax, large exposure differences, as well as geometric distortion.
To minimize the appearance change between the three views and to represent the wide FOV of the stitched frames, we first adopt a camera pose transformation to warp $\ci_L$ and $\ci_R$ to the position of $\co_L$ and $\co_R$, respectively.
Therefore, the new origin is set at the center camera $C_M$.
Then, we apply a cylindrical projection (by approximating the scene to be at infinity) to warp all the views onto a common viewing cylinder, as shown in Figure~\ref{fig:camera}(a).
However, even after camera calibration, exposure compensation, fisheye distortion correction, and cylindrical projection, parallax still causes significant misalignment, which results in severe ghosting artifacts, as shown in Figure~\ref{fig:cylinder}(d).

\begin{figure*}
	\centering
	\footnotesize
	\includegraphics[width=1.0\textwidth]{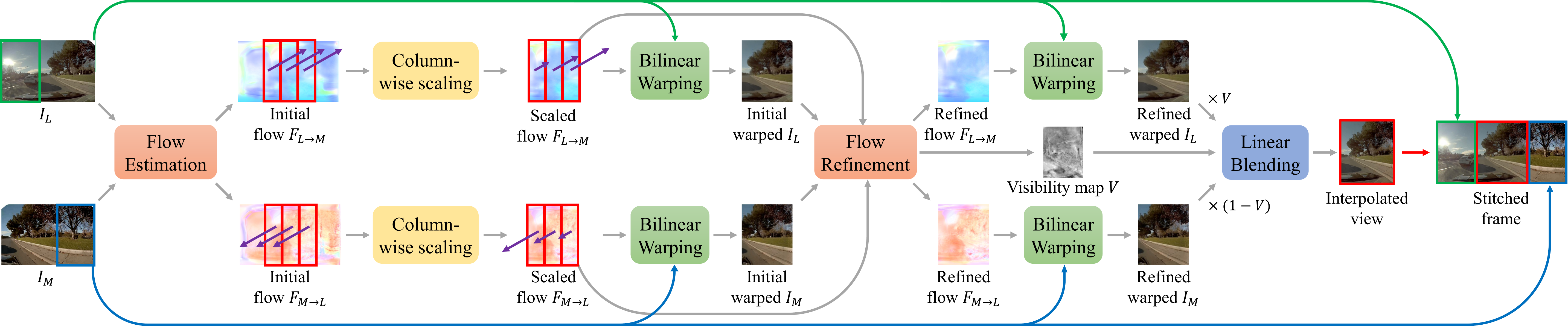}
	\\
	\caption{
		\textbf{Pushbroom interpolation layer.}
		A straightforward implementation of the pushbroom interpolation layer requires to generate all the intermediate flows and the intermediate views, which is time-consuming when the number of interpolated views $K$ is large.
		Therefore, we develop a fast pushbroom interpolation layer by a column-wise scaling on optical flows, which only requires to generate one interpolated image for any given $K$.
	}
	\label{fig:pushbroom}
\end{figure*}

\subsection{Formulation}
In this work, we cast video stitching as a problem of spatial interpolation between the side views and the center view.
We denote the output view by $\mathcal{O}$, and the input views (projected onto the output cylinder) by $I_L$, $I_M$, and $I_R$, respectively.
Note that $I_L$, $I_M$, and $I_R$ are in the same coordinate system and have the same resolution.
We define a \emph{transition region} as part of the overlapping region between a pair of inputs (see the yellow regions in Figure~\ref{fig:camera}(b)).
Within the transition region, we progressively warp $K$ vertical slices from both images to create a smooth transition from one camera to another.
Outside the transition region, we directly take the pixel values from the input images without modifying them.

For presentation clarity, here we focus only on $I_L$ and $I_M$.
Our goal is to generate $K$ intermediate frames, $\hat{I}_L^{(k)}$, which smoothly transition between $I_L$ and $I_M$.
We first compute the bidirectional optical flows, $F_{L\rightarrow M}$ and $F_{M\rightarrow L}$, and then generate warped frames
$\hat{I}_{L}^{(k)} = \mathcal{W}(I_L, \alpha_k \cdot F_{L\rightarrow M})$ and $\hat{I}_{M}^{(k)} = \mathcal{W}(I_M, (1-\alpha_k) \cdot F_{M\rightarrow L})$, where $\mathcal{W}(I, F)$ is a function that warps image $I$ based on flow $F$, and $\alpha_k=\{k/K\}_{k=1,...,K}$ scales the flow to create the smooth transition.
We define the left stitching boundary $b_L$ as the column of the leftmost valid pixel for $I_M$ on the output cylinder.
Given the interpolation step size $s$, the left half of the output view, $\mathcal{O}_L$, is constructed by
\begin{align}
	\mathcal{O}_L(x) \!=\! 
	\begin{cases}
		I_L(x), & \text{if}~0 \leq x < b_L \\
		\hat{I}_{LM}^{(k)}(x), & \text{if}~b_L \!+\! (k\!-\!1) \!\cdot\! s \!\leq\! x \!<\! b_L \!+\! k \!\cdot\! s \\
		I_M(x), & \text{if}~b_L \!+\! K \!\cdot\! s \!\leq\! x \!<\!  \frac{W}{2},
	\end{cases}
	\label{eq:pushbroom-left}
\end{align}
where $W$ is the width of the output frame, and $\hat{I}_{LM}^{(k)}$ is obtained by appropriately fusing $\hat{I}_{L}^{(k)}$ and $\hat{I}_{M}^{(k)}$, $k\!=\!1,\!...\!,K$ (see Section~\ref{sec:pushbroom_layer}). 
By construction, the output image is aligned with $I_L$ at $b_L$, and aligned with $I_M$ at $b_L + K \cdot s$.
Within the transition region, the output view gradually changes from $I_L$ to $I_M$ by taking the corresponding columns from the intermediate views.
The right half part of the output, $\mathcal{O}_R$, is defined similarly to $\mathcal{O}_L$. 
Figure~\ref{fig:cylinder}(e) shows a result of this interpolation.

We note that the finer the interpolation steps, the higher the quality of the stitched results. 
We empirically set $K = 100$ and $s = 2$, \ie, $100$ pushbroom columns each $2$-pixel wide.

\subsection{Fast Pushbroom Interpolation Layer}
\label{sec:pushbroom_layer}
Synthesizing the transition regions exactly as described in the previous section is computationally expensive.
For each side, it requires scaling the forward and backward optical flow fields $K$ times, and using them to warp the full-resolution images just as many times. 
For $H\times W$ images, this results in $2 \times H\times W \times K$ pixels to warp for each side.
However, we only need a slice of $s=2$ pixels from each of them.

Instead of scaling each flow field in its entirety, we propose to generate a single flow field in which entries corresponding to different slices are scaled differently.
For instance, from the flow field from $I_L$ to $I_M$, we generate a new field 
\begin{align}
	\hat{F}_{L \!\rightarrow\! M}(x) \!=\! 
	\begin{cases}
		0, & \text{if}~0 \leq x < b_L \\
		\alpha_k F_{L \!\rightarrow\! M}(x), & \text{if}~ b^{(k)} \!\leq\! x \!<\! b^{(k+1)} \\
		F_{L \!\rightarrow\! M}(x), & \text{if}~b_L \!+\! K \!\cdot\! s \!\leq\! x \!<\! \frac{W}{2},
	\end{cases}
	\label{eq:scaled-flow-left}
\end{align}
where $b^{(k)} = b_L + (k - 1) \cdot s$ are the boundaries of each slice.
We can then warp both images as $\hat{I}_{L} = \mathcal{W}(I_L, \hat{F}_{L\rightarrow M})$ and
$\hat{I}_{M} = \mathcal{W}(I_M, \hat{F}_{M\rightarrow L})$, where $\hat{F}_{M\rightarrow L}$ is computed with Equation~\ref{eq:scaled-flow-left} with $(1-\alpha_k)$ in place of $\alpha_k$.
Note that this approach only warps each pixel in the input images \emph{once}.

To deal with the unavoidable artifacts of optical flow estimation, we use a flow refinement network to refine the scaled flows and predict a visibility map for blending.
As shown in Figure~\ref{fig:pushbroom}, the flow refinement network takes the scaled optical flows and the initial estimates of the warped images, from which it generates refined flows and a visibility map $V$.
The visibility map can be considered as a quality measure of the flow, which prevents any potential ghosting artifacts due to occlusions. 
With the refined flows, we warp the input images again to obtain $\tilde{I}_{L}$ and $ \tilde{I}_{M}$.
The final interpolated image is then generated by blending based on visibility: $\tilde{I}_{LM} = V \cdot \tilde{I}_{L} + (1-V) \cdot \tilde{I}_{M}$.

Finally, the output view, $\mathcal{O}_L$, is constructed by replacing all the $\hat{I}_{LM}^{(k)}$ in Equation~\ref{eq:pushbroom-left} with $\tilde{I}_{LM}$.
We generate $\tilde{I}_{RM}$ and construct $\mathcal{O}_R$ by mirroring the process above.
Our fast pushbroom interpolation layer generates the results with similar quality but is about $40\times$ faster than the direct implementation for an output image with a resolution of $1000 \times 600$ pixels.

\subsection{Training Pushbroom Stitching Network}
\label{sec:implementation}

\paragraph{Training dataset.}
Capturing data to train our system is challenging, as one would need to use hundreds of synchronized linear cameras.
Instead, we render realistic synthetic data using the urban driving simulator CARLA~\cite{carla}, which allows us to specify the location and rotation of cameras.
For the input cameras, we follow the setup of Figure~\ref{fig:camera}(a).
To synthesize the output pushbroom camera, we use 100 cameras uniformly spaced between $C^{i}_{L}$ and $C_{M}$, and between $C^{i}_{R}$ and $C_{M}$.
We then use Equation~\ref{eq:pushbroom-left} and replace $I_{LM}^{(k)}$ with these views to render ground-truth stitched video.
We synthesize 152 such videos with different routes and weather conditions (\eg, sunny, rainy, cloudy, etc.) for training.
We provide the detailed network architectures of the flow estimation and flow refinement networks in the supplementary material.

\paragraph{Training loss.}
To train our pushbroom interpolation network, we optimize the following loss functions: (1) content loss, (2) perceptual loss, and (3) temporal warping loss.

The content loss is computed by $\mathcal{L}_C = \sum_{x,y} M_{x,y} \cdot \| \mathcal{O}_{x,y} - \mathcal{S}_{x,y} \|_1$, where $\mathcal{O}$ is the output image, $\mathcal{S}$ is the ground-truth, and $M_{x,y}$ is a mask indicating whether pixel $x,y$ is valid on the viewing cylinder.
The perceptual loss is computed by $\mathcal{L}_P = \sum_{x,y} \tilde{M}_{x,y} \cdot \| \phi_{x,y}(\mathcal{O}) - \phi_{x,y}(\mathcal{S}) \|_1$, 
where $\phi(\cdot)$ denotes the feature activation at the \texttt{relu4-3} layer of the pre-trained VGG-19 network~\cite{VGG} and $\tilde{M}$ is the valid mask downscaled to the size of the corresponding features.
To improve the temporal stability, we also optimize the temporal warping loss~\cite{Lai-ECCV-2018} $\mathcal{L}_{T} = \sum_{t' \in \Omega_t} \sum_{x,y} M_{x,y} \cdot C^{t \Rightarrow t'}_{x,y} \cdot \| \mathcal{O}^{(t)}_{x,y} - \hat{\mathcal{O}}^{(t')}_{x,y} \|_1$, 
where $\Omega_t$ is the set of neighboring frames at time $t$, $C$ is a confidence map, and $\hat{\mathcal{O}}^{(t')} = \mathcal{W}(\mathcal{O}^{(t')}, F^{t \Rightarrow t'})$ is the frame warped with optical flow $F^{t \Rightarrow t'}$.
We use PWC-Net~\cite{sun2018pwc} to compute the optical flow between subsequent frames.
Note that the optical flow $F^{t \Rightarrow t'}$ is only used to compute the training loss, and is not needed at testing time.
The confidence map $C^{t \Rightarrow t'} = \exp( -\alpha \| \mathcal{S}^{(t)} - \hat{\mathcal{S}}^{(t')} \|_2^2 )$ is computed from the ground-truth frame $\mathcal{S}^{(t)}$ and $\mathcal{S}^{(t')}$, where $C \in [0, 1]$.
A smaller value of $C$ indicates that the pixel is more likely to be occluded.

The overall loss function is defined as $\mathcal{L} = \lambda_C \mathcal{L}_C + \lambda_P \mathcal{L}_P + \lambda_T \mathcal{L}_T$, 
where $\lambda_C$, $\lambda_P$, and $\lambda_T$ are balancing weights set empirically.
We empirically set $\lambda_C = 1$, $\lambda_P = 0.001$, and $\lambda_T = 10$.
For the spatial optical flow in the transition regions, we use SuperSloMo~\cite{jiang2018super} initialized with the weights provided by the authors and then fine-tuned to our use-case in our end-to-end training.
We provide more implementation details in the supplementary material.

\section{Experimental Results}
\label{sec:experiment}

\begin{figure}
	\centering
	\subfloat[Stitched frame ($\mathcal{L}_c$+$\mathcal{L}_p$+$\mathcal{L}_t$)]{\includegraphics[height=0.725in]{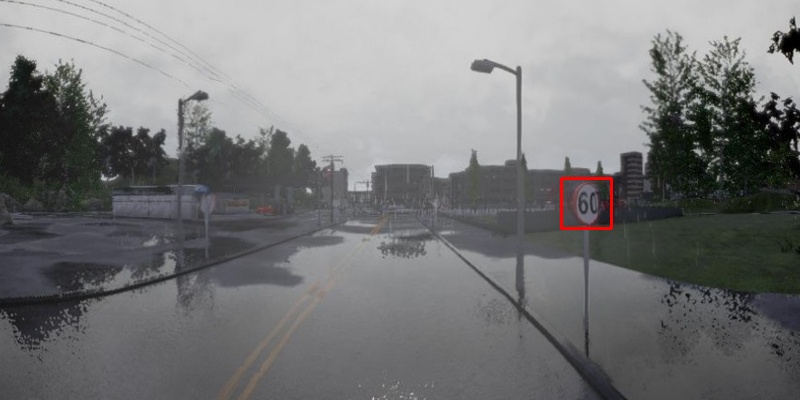} }
	\subfloat[GT]{\includegraphics[height=0.725in]{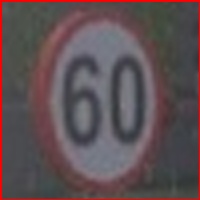}}
	\subfloat[Baseline]{\includegraphics[height=0.725in]{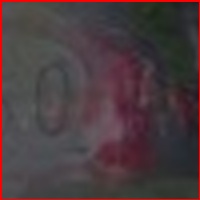}}
	\subfloat[$\mathcal{L}_c$]{\includegraphics[height=0.725in]{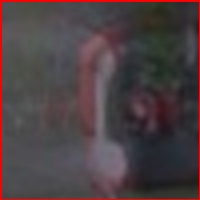}}
	\subfloat[$\mathcal{L}_c$+$\mathcal{L}_p$]{\includegraphics[height=0.725in]{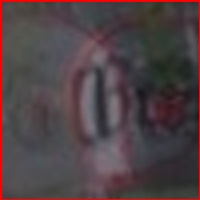}}
	\subfloat[$\mathcal{L}_c$+$\mathcal{L}_p$+$\mathcal{L}_t$]{\includegraphics[height=0.725in]{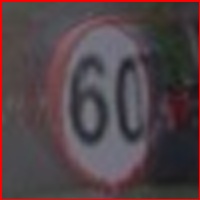}}
	\vspace{2mm}
	\caption{
		\textbf{Stitching on a synthetic video.}
		After training the proposed model on the synthetic data, our model aligns the content well and reduces ghosting artifacts.}\label{fig:ablation}
\end{figure}

The output of our algorithm, while visually pleasing, does not match a physical optical system, since the effective projection matrix changes horizontally.
To numerically evaluate our results we can use rendered data (Section~\ref{sec:analysis}).
However, a pixel-level numerical comparison with other methods is not possible as each method effectively uses a different projection matrix.
For a fair comparison, then, we carried out an extensive user study (Section~\ref{sec:comparison}).
The video results are available in the supplementary material and our project website at \url{http://vllab.ucmerced.edu/wlai24/video_stitching/}.

\subsection{Model Analysis}\label{sec:analysis}
To quantitatively evaluate the performance of the stitching quality, we use the CARLA simulator to render a test set using a different town map from the training data.
We render 10 test videos, where each video has 300 frames.

We measure the PSNR and SSIM~\cite{wang2004image} between the stitched frames and the ground-truth images for evaluating the image quality.
In addition, we measure the temporal stability by computing the temporal warping error $E_{\text{warp}} = \sum_{t=1}^{T-1} \frac{1}{| \tilde{M}^{(t)} |} \sum_{x,y \in \tilde{M}^{(t)}} \| O_{x,y}^{(t)} - \hat{O}_{x,y}^{(t+1)} \|_2^2$,
where $\hat{O}^{(t+1)}$ is the flow-warped frame $O^{(t+1)}$, $\tilde{M}^{(t)}$ is a mask indicating the non-occluded pixels, and $| \tilde{M}^{(t)} |$ is the number of valid pixels in the mask.
We use the occlusion detection method by Ruder~\etal~\cite{Ruder-2016} to estimate the mask $\tilde{M}^{(t)}$.

We first evaluate the baseline model, where the pushbroom interpolation layer is initialized with the pre-trained SuperSloMo~\cite{jiang2018super}.
The baseline model provides a visually plausible stitching result but causes object distortion and temporal flickering due to inaccurate flow estimation.
After fine-tuning the whole model, both the visual quality and temporal stability are significantly improved.
As shown in Table~\ref{tab:ablation}, all the loss functions, $\mathcal{L}_C$, $\mathcal{L}_P$, and $\mathcal{L}_T$, improve the PSNR and SSIM and also reduce the temporal warping error.
In Figure~\ref{fig:ablation}, we show an example where our full model aligns the speed sign well and avoids the ghosting artifacts.

\begin{figure}
	\centering
	\footnotesize
	\renewcommand{\tabcolsep}{2pt} 
	\begin{tabular}{ccc}
		\includegraphics[height=0.20\columnwidth]{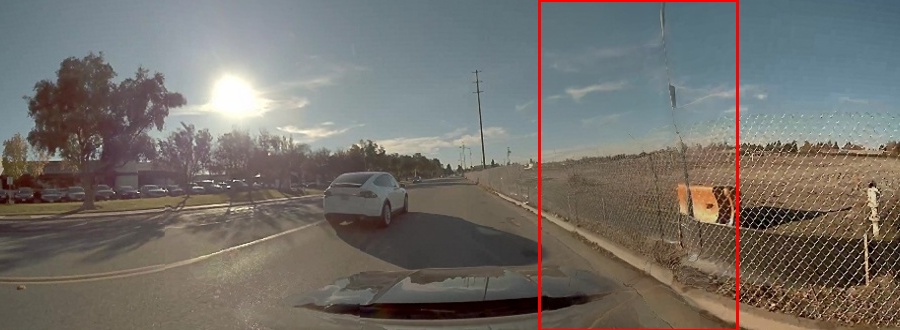}
		&
		\includegraphics[height=0.20\columnwidth]{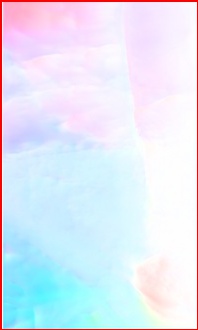}
		&
		\includegraphics[height=0.20\columnwidth]{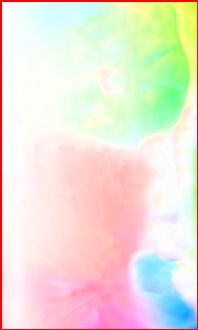}
		\\
		\includegraphics[height=0.20\columnwidth]{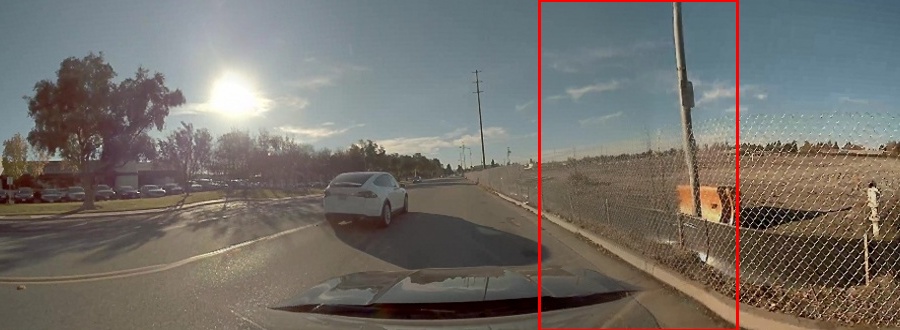}
		&
		\includegraphics[height=0.20\columnwidth]{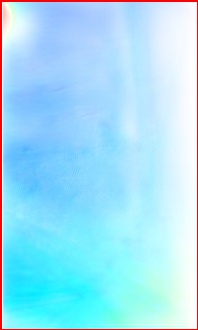}
		&
		\includegraphics[height=0.20\columnwidth]{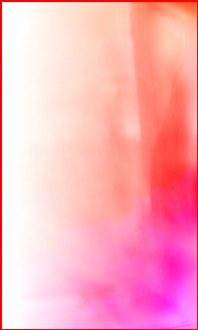}
		\\
		(a) & (b) & (c)
	\end{tabular}
	\vspace{2mm}
	\caption{
		\textbf{Visualization of the stitched frames and flows.}
		We show the stitched frames (a), forward flows (b), and backward flows (c) from the pushbroom interpolation layer before (top) and after (bottom) fine-tuning the proposed model.
		The fine-tuned model generates smooth flow fields to warp the input views and preserve the content (e.g., the pole on the right) well.
	}
	\label{fig:visualization}
\end{figure}

Figure~\ref{fig:visualization}(a) shows a stitched frame from the baseline model, where the pole on the right is distorted and almost disappears.
After training, the pole remains intact, Figure~\ref{fig:visualization}(b).
We also visualize the optical flows before and after training the model.
After end-to-end training, the flows are smoother and warp the pole as a whole, avoiding distortion.

\begin{figure}
	\begin{minipage}{0.48\textwidth}
		\scriptsize
		\centering
		\begin{tabular}{lccc}
			\toprule
			Training loss & 
			PSNR $\uparrow$ & 
			SSIM $\uparrow$ & 
			$E_{\text{warp}}$ $\downarrow$ \\
			\midrule
			N.A. (baseline)
			& 27.69 & 0.908 & 13.89 $\times 10^{-4}$ \\
			$\mathcal{L}_C$ 
			& 30.72 & 0.925 & 11.72 $\times 10^{-4}$\\
			$\mathcal{L}_C$ + $\mathcal{L}_P$ 
			& 31.04 & 0.926 & 11.63 $\times 10^{-4}$\\
			$\mathcal{L}_C$ + $\mathcal{L}_P$ + $\mathcal{L}_T$ 
			& 31.27 & 0.928 & 10.67 $\times 10^{-4}$ \\
			\bottomrule
		\end{tabular}
		\vspace{3mm}
		\captionof{table}{
			\textbf{Ablation study.}
			After training the model with the content loss $\mathcal{L}_C$, perceptual loss $\mathcal{L}_P$, and the temporal loss $\mathcal{L}_T$, the image quality and temporal stability are significantly improved.	
		}
		\label{tab:ablation}
	\end{minipage}
	\hfill
	\begin{minipage}{0.48\textwidth}
		\scriptsize
		\centering
		\renewcommand{\tabcolsep}{4pt} 
		\begin{tabular}{c|cccc}
			\toprule
			\multirow{2}{*}{Ours vs.} &
			\multirow{2}{*}{Preference} & 
			Broken & 
			Less & 
			Similar 
			\\
			& & objects & ghosting & results \\
			\midrule
			AutoPano~\cite{autopano} &
			90.74$\%$ & 85.71$\%$ & 20.41$\%$ & 10.20$\%$ \\
			VRWorks~\cite{vrworks} &
			97.22$\%$ & 80.00$\%$ & 49.52$\%$ & 1.90$\%$ \\
			STCPW~\cite{jiang2015video} &
			98.15$\%$ & 87.74$\%$ & 38.68$\%$ & 0$\%$ \\
			\midrule
			Overall &
			95.37$\%$ & 84.74$\%$ & 36.57$\%$ & 3.88$\%$ \\ 
			\bottomrule
		\end{tabular}
		\vspace{1mm}
		\captionof{table}{\textbf{User study.}
			We conduct pairwise comparisons on 20 real videos.
			Our method is preferred by $95\%$ of users on average.}
		\label{tab:userstudy}
	\end{minipage}
	
\end{figure}

\begin{figure}
	\centering
	\subfloat[STCPW~\cite{jiang2015video}]{
		\includegraphics[width=0.49\textwidth]{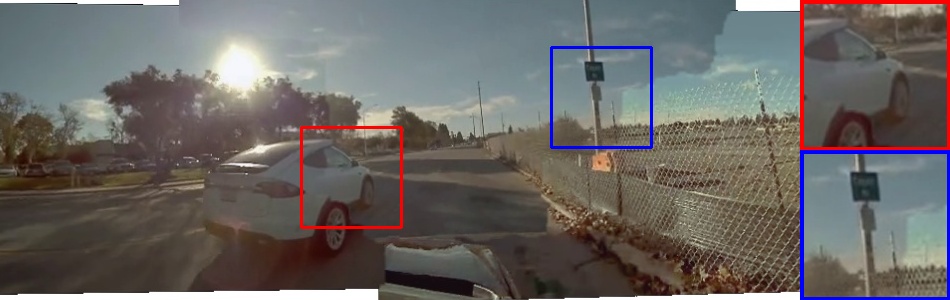}
		\includegraphics[width=0.49\textwidth]{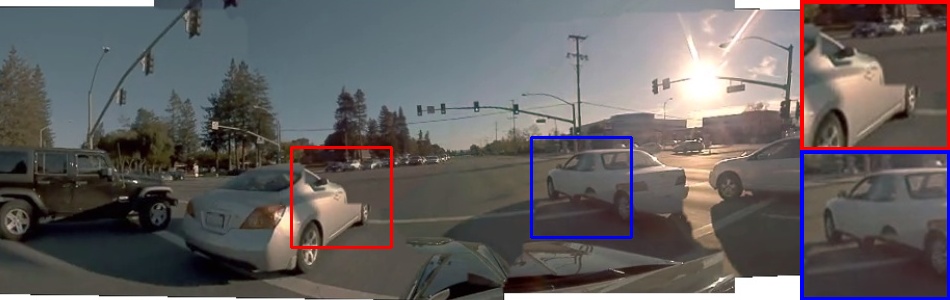}}
	\vspace{-3mm}
	\\
	\subfloat[AutoPano Video~\cite{autopano}]{\includegraphics[width=0.49\textwidth]{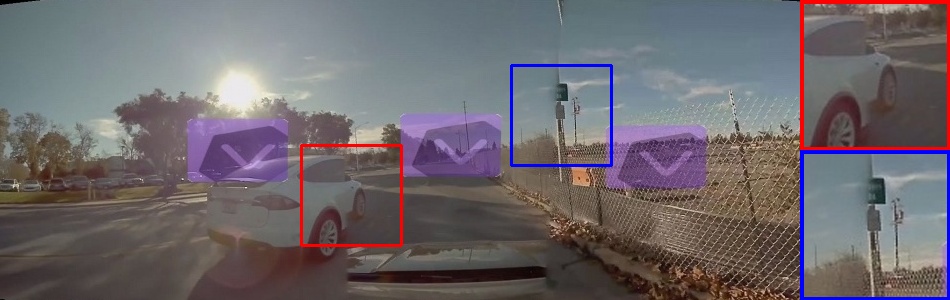}
		\includegraphics[width=0.49\textwidth]{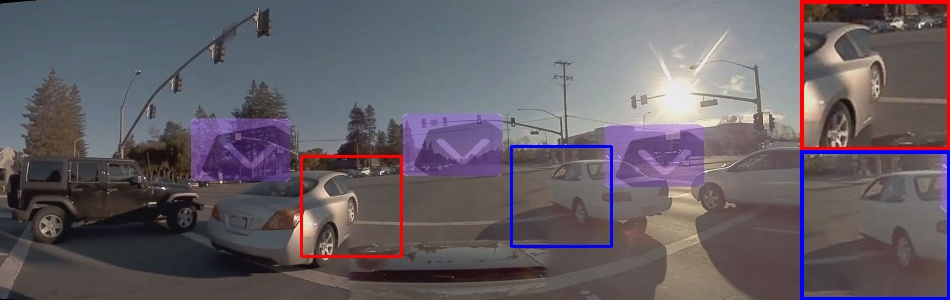}}
	\vspace{-3mm}
	\\
	\subfloat[NVIDIA VRWorks~\cite{vrworks}]{\includegraphics[width=0.49\textwidth]{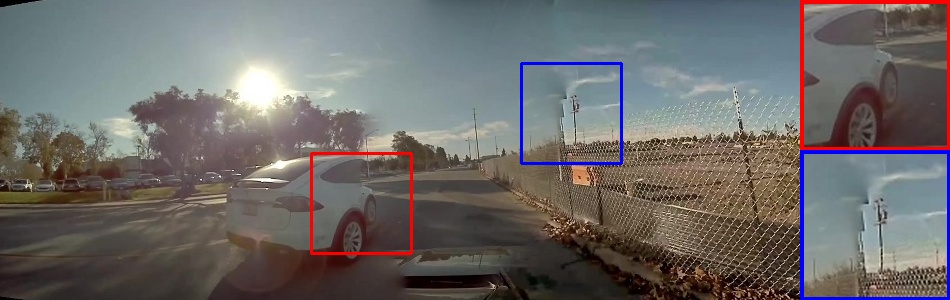}
		\includegraphics[width=0.49\textwidth]{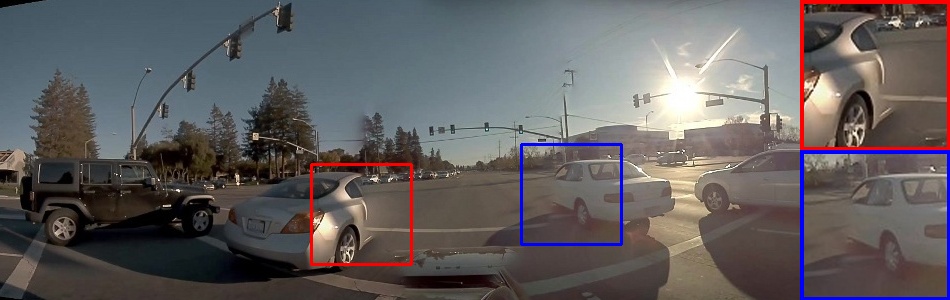}}
	\vspace{-3mm}
	\\
	\subfloat[Ours]{\includegraphics[width=0.49\textwidth]{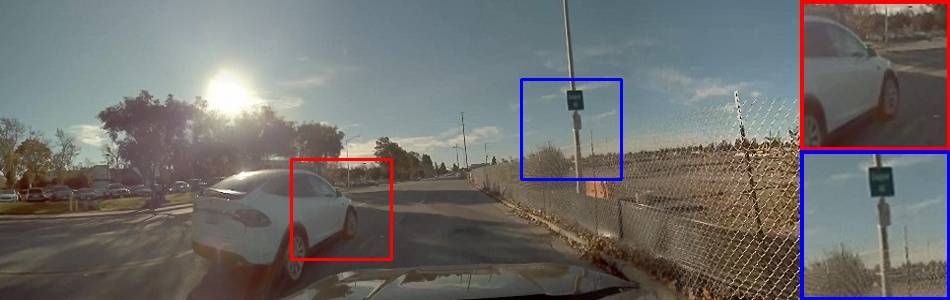}
		\includegraphics[width=0.49\textwidth]{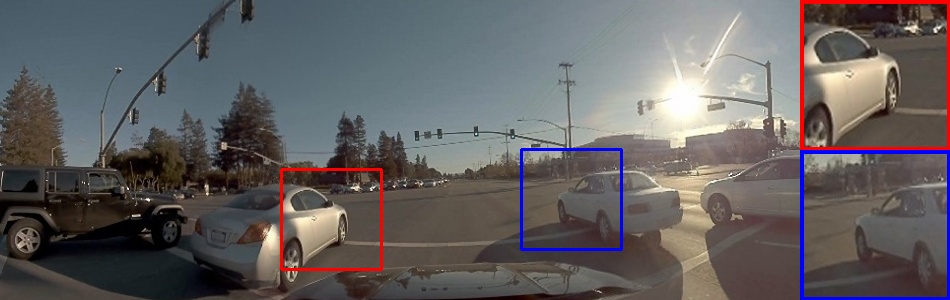}}
	\vspace{2mm}
	\caption{
		\textbf{Comparison with existing video stitching methods.}
		The proposed method achieves better alignment quality while better preserving the shape of objects and avoiding ghosting artifacts.
	}
	\label{fig:compare_stoa}
\end{figure}

\subsection{Comparisons with Existing Methods}\label{sec:comparison}

We compare the proposed method with commercial software, AutoPanoVideo~\cite{autopano}, and existing video stitching algorithms, STCPW~\cite{jiang2015video} and NVIDIA VRWorks~\cite{vrworks}.
We show two stitched frames from real videos in Figure~\ref{fig:compare_stoa}, where the proposed approach generally achieves better alignment quality with fewer broken objects and ghosting artifacts.
More video results are provided in the supplementary material.

As different methods use different projection models, a fair quantitative evaluation of the different video stitching algorithms is impossible.
Therefore, we conduct a human subject study through pairwise comparisons.
Specifically, we ask the participants to indicate which stitched video presents fewer artifacts from a pair of videos.
We evaluate a total of 20 real videos and ask each participant to compare 12 pairs of videos.
In each comparison, participants can watch both videos for multiple times before making a selection.
In total, we collect the results from 54 participants.

Table~\ref{tab:userstudy} shows that our results are preferred by about $95\%$ of users, which demonstrates the effectiveness of the proposed method on generating high-quality stitching results.
In addition, we ask participants to provide the reasons why they prefer the selected video from the following options:
(1) the video has fewer broken lines or objects, 
(2) the video has less ghosting artifacts,
and (3) the two videos are similar.
Overall, our results are preferred due to better alignment and fewer broken objects.
Moreover, only $4\%$ of users feel that our result is comparable to the others, which indicates that users generally have a clear judgment when comparing our method with other approaches.

\subsection{Discussion and Limitations}

Our method requires the cameras to be calibrated for the cylindrical projection of the inputs.
While common to many stitching methods, \eg, NVIDIA's VRWorks~\cite{vrworks}, this requirement can be limiting, if strict.
However, our experiments reveal that moving the side cameras inwards by up to $62.5\%$ of the original baseline, reduces the PSNR by less than 1dB.
An outward shift is more problematic because it reduces the overlap between the views.
Still, an outward shift that is $30\%$ of the original baseline causes less than 2dB drop in PSNR.
Fine-tuning the network by perturbing the original configuration of cameras can reduce the error.
We present detailed analysis in the supplementary material.
Our method inherits some limitations of the optical flow.
For instance, thin structures can cause a small amount of ghosting effects.
We show failure cases in the supplementary material.
In practice, the proposed method performs robustly even in such cases.

\section{Conclusion}
\label{sec:conclusion}

In this work, we present an efficient algorithm to stitch videos with deep CNNs.
We propose to cast video stitching as a problem of spatial interpolation and we design a pushbroom interpolation layer for this purpose.
Our model effectively aligns and stitches different views while preserving the shape of objects and avoiding ghosting artifacts.
To the best of our knowledge, ours is the first learning-based video stitching algorithm.
A human subject study demonstrates that it outperforms existing algorithms and commercial software.

\bibliography{reference}

\end{document}